\documentclass[a4paper, 10pt, conference]{ieeeconf}      
\usepackage{FG2021}

\FGfinalcopy 

\IEEEoverridecommandlockouts                              
\usepackage{graphics} 
\usepackage{epsfig} 
\usepackage{mathptmx} 
\usepackage{times} 
\usepackage{amsmath} 
\usepackage{amssymb}  

\usepackage{mathtools, stmaryrd}

\newcommand{\overbar}[1]{\mkern 1.5mu\overline{\mkern-1.5mu#1\mkern-1.5mu}\mkern 1.5mu}

\DeclareMathOperator*{\argmin}{arg\,min} 

\def\FGPaperID{305} 

\title{\LARGE \bf
Monocular Human Shape and Pose\\ 
with Dense Mesh-borne Local Image Features}

\author{\parbox{16cm}{\centering
    {\large Shubhendu Jena, Franck Multon, Adnane Boukhayma}\\
    {\normalsize Inria, Univ. Rennes, CNRS, IRISA, M2S, France
    \\}}
}

\begin{document}

\ifFGfinal
\thispagestyle{empty}
\pagestyle{empty}
\else
\author{Anonymous FG2021 submission\\ Paper ID \FGPaperID \\}
\pagestyle{plain}
\fi
\maketitle

\begin{abstract}

We propose to improve on graph convolution based approaches for human shape and pose estimation from monocular input, using pixel-aligned local image features. Given a single input color image, existing graph convolutional network (GCN) based techniques for human shape and pose estimation (e.g. \cite{kolotouros2019convolutional}) use a single convolutional neural network (CNN) generated global image feature appended to all mesh vertices equally to initialize the GCN stage, which transforms a template T-posed mesh into the target pose. In contrast, we propose for the first time the idea of using local image features per vertex. These features are sampled from the CNN image feature maps by utilizing pixel-to-mesh correspondences generated with DensePose \cite{guler2018densepose}. Our quantitative and qualitative results on standard benchmarks show that using local features improves on global ones and leads to competitive performances with respect to the state-of-the-art.    

\end{abstract}

\section{INTRODUCTION}

Reconstructing human bodies e.g. \cite{moon2020i2l,lin2021end} and their parts (faces e.g. \cite{feng2018joint,abrevaya2020cross}, hands e.g. \cite{boukhayma20193d,ge20193d}) from minimal, partial and noisy inputs is one of the most sought-after goals of human centered machine learning.  In this regard, human shape and pose recovery from a single color image remains a popular problem in computer vision and graphics spurring a vast research literature, with applications in various areas such as action recognition, avatarization, human machine interaction, etc.
While earlier approaches to 3D reconstruction relied on multi-view triangulation or depth information, the recent surge of deep learning has allowed the reduction of acquisition constraints to as little as a single input RGB image. The ill-posedness of this monocular  setting is alleviated through learning strong statistical priors from large training datatsets with deep CNNs, which has shown to be successful especially for single shape class settings such as humans. Such approaches also benefit from transfer learning techniques by leveraging networks pre-trained on massive general datasets (e.g. ImageNet \cite{deng2009imagenet}). While several work \cite{kanazawa2018end,kolotouros2019learning,boukhayma20193d} show that the learning can be further regularized by integrating differentiable parametric naked human body models (e.g. SMPL \cite{loper2015smpl}, GHUM \cite{xu2020ghum}, Frank \cite{joo2018total}) within deep networks, other methods advocate predicting model-free 3D shape outputs (e.g. \cite{moon2020i2l,kolotouros2019convolutional}), with the prospect of more expressive results, whilst the parametric model is usually involved in generating training 3D pseudo-ground-truth in this case.

In this work, given a single color image, we tackle the problem of  estimating human 3D shape and pose in the form of a fixed-topology triangle mesh, using a feed forward deep neural network. We specifically focus on improving on a class of model-free methods that propose to use GCNs for this task \cite{kolotouros2019convolutional,ge20193d,kulon2020weakly,kulon2019single}, the graph's vertices and edges being defined as those of the 3D mesh representing the human shape. Traditionally, these methods extract a global latent feature vector from the image using a CNN, and this same vector is used as input feature to all the mesh vertices equally, as in \cite{kolotouros2019convolutional}. The GCN then starts from these mesh-borne features and deforms a T-posed template mesh towards the target posed mesh. A noteworthy variant of this strategy consists in predicting a global feature and mapping it subsequently to initial low resolution mesh vertex features \cite{ge20193d,kulon2020weakly,kulon2019single}, and it was mostly explored for 3D hand prediction rather than full body partly due to the smaller mesh size.

\begin{figure*}[t]
\centering
\includegraphics[width=0.9\linewidth]{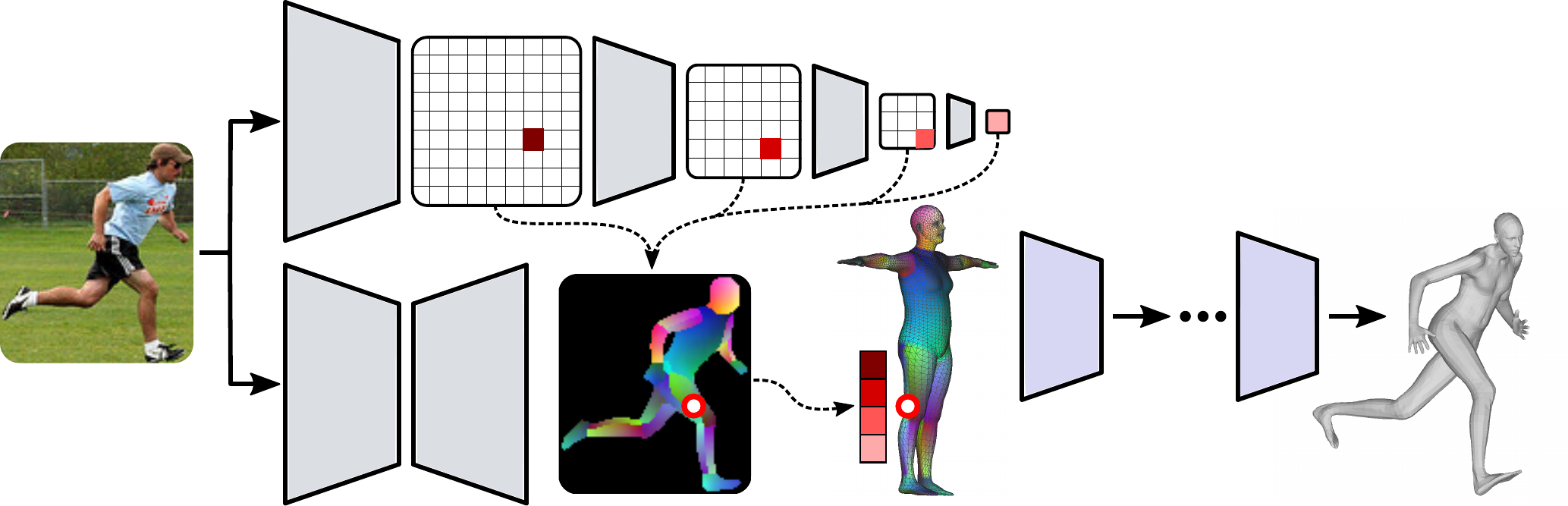}
\caption{Overview.
Given an input color image, DensePose \cite{guler2018densepose} \textit{(Bottom left)} produces dense pixel-to-surface correspondences. Meanwhile, an image convolutional neural network (CNN) \textit{(Top)} builds feature maps at multiple depths \textit{(Shades of red)}. The correspondences are then used to sample \textit{(Dashed lines)} local image features from the CNN feature maps for each template surface vertex at its corresponding image location \textit{(Red Circle)}. Next, we use a graph convolutional network (GCN) \textit{(Right)} to map the template surface with vertex specific local image features to the final posed surface.     
}
\label{fig:pipe}
\end{figure*}

In contrast to these methods, we propose here to use per vertex pixel-aligned local image features as initialization for the GCN stage, as illustrated in Fig.\ref{fig:pipe}. We use a method for predicting dense pixel-to-surface correspondences (DensePose \cite{guler2018densepose}) of humans to map each visible vertex in the template geometry to a pixel in the input image. Bi-linear interpolation is then use to build a different feature vector per vertex, by sampling and stacking local image features at the vertex's corresponding 2D location in the image space, at different CNN feature depths. Given a T-posed initial mesh appended with vertex specific local image features, a GCN regresses the final mesh vertex positions. The network composed of the image CNN and mesh GCN is trained end-to-end using 3D supervision following the training scheme in \cite{moon2020i2l}. 

We evaluate our method on standard benchmarks for human mesh prediction from images, namely 3DPW \cite{von2018recovering} and Human3.6M \cite{ionescu2013human3}. Our numerical and visual results demonstrate that using local features improves on using only global ones in GCN based human mesh recovery from single image. We also show that our method yields competitive results in comparison to the state-of-the-art methods.    

\section{Related work}
There is a substantial body of work on the subject of human shape and pose estimation from a single image. We review in this section works that we deemed most relevant to the context of our contribution.

\subsection{Optimization-based methods}
Most current optimization-based methods rely on using 2D landmarks such as key-points and silhouettes~\cite{sigal2007combined,bogo2016keep,guan2009estimating} and optimize parametric models such as SCAPE~\cite{anguelov2005scape} and SMPL~\cite{loper2015smpl} to fit them to these landmarks. The optimization objective consists of mainly two kinds of loss terms. The first is prior terms designed to penalize unnatural human poses and shapes. The second is data terms minimizing the difference between the inferred 2D landmarks obtained by projecting the predicted mesh onto the image and the ground-truth body landmarks. Additionally, there have been other recent methods~\cite{zanfir2018monocular} which incorporated additional terms such as part segmentation, scene and temporal constraints in the optimization objective. Although optimization-based methods provide generally reliable solutions, they are notoriously slow and prone to getting stuck in local minima especially with challenging initializations. This incentivizes learning-based solutions such as ours, which offer faster inference and do not require initialization.

\subsection{Learning-based methods}

\subsubsection{Model-based}
Model-based methods make use of a parametric 3D human body model to perform 3D human pose and shape estimation. The learning problem is thus reduced to learning the parameters associated with the body model from images and other types of input. A notable example is the work of~\cite{kanazawa2018end} which directly regresses the SMPL parameters from a single RGB image, by performing a weak supervision comprising of a 2D key-point reprojection error and an adversarially learnt pose prior. SPIN~\cite{kolotouros2019learning} improves on this with a self-improving framework that incorporates 3D human model parameter optimization into the network training process. To deal with occlusions and noisy situations which make image-based methods more susceptible to failure, some approaches use alternative inputs such as 2D joint heatmaps~\cite{tung2017self}, silhouettes~\cite{pavlakos2018learning,varol2018bodynet} and semantic segmentation maps~\cite{omran2018neural}. In-spite of the aforementioned advantages behind model-based methods, such strategies can conversely also be somewhat restrictive. The tight relationship between the parameters and the model's output reduces the expressiveness of the generated meshes. Hence, as a model-free method, our framework focuses instead on directly regressing the 3D human body mesh vertices corresponding to an input image following seminal work.

\subsubsection{Model-free}
As the name implies, model-free methods do not rely on parametric models for 3D human body reconstruction. Instead, they directly regress an explicit body shape representation from the input images. Some of the earlier work uses a volumetric reconstruction approach with a voxel output \cite{varol2018bodynet, zheng2019deephuman}. The main drawback of voxel-based methods is their inability to represent detailed surfaces due to memory limitations. Other methods use different representations such as depth maps, point clouds, etc. \cite{gabeur2019moulding}. However, these suffer mostly from lacking surface continuity and neighborhood connectivity. To deal with these problems, recent methods propose to directly regress 3D SMPL meshes representing the output human body shape. To the best of our knowledge, this line of work started with CMR~\cite{kolotouros2019convolutional} which used a GCN to directly regress 3D coordinates of vertices from a global image feature. Pose2Mesh~\cite{choi2020pose2mesh} also did the same but from 2D joints as input instead. The work of Lin et al. \cite{lin2021end} yields arguably state-of-the-art performances through the use of Transformers \cite{vaswani2017attention} but requires considerable training time and data. Furthermore, there has been another class of methods focusing on learning dense correspondence between 2D images and 3D shapes. The seminal work of 
DensePose~\cite{guler2018densepose} which provides dense mapping from images to a human body model by regressing 2D correspondence maps has led arguably to the advent of much similar work focusing on dense correspondence. \cite{xu2019denserac,rong2019delving} utilized DensePose correspondence maps for 3D human model recovery. However, they only leverage them as input images. Zhang et al.~\cite{zhang2020learning} predict and use local and global correspondence maps as input to further CNN stages for pose and shape prediction in a similar fashion. In contrast, we propose here to use correspondence maps to provide vertices with semantically meaningful image-aligned local features. In fact, our strategy is similar to DecoMR~\cite{zeng20203d}, where the authors establish dense correspondences between the surface and the input image, which is subsequently used to transfer image features to the UV-map domain and thereafter perform 3D coordinate regression with a 2D CNN. Differently, we explore here the idea of performing vertex position regression from local image features using a convolutional network defined naturally on the same data representation as the output (i.e. GCN on the mesh as opposed to CNN on the UV-map). 

\section{METHOD}

We describe in this section the various components of our method, which follows the illustration in Fig.\ref{fig:pipe}. Our input is a single RGB image and our output is a triangle human mesh with the SMPL \cite{loper2015smpl} template topology. Given the input image, a convolutional neural network (CNN) is tasked with extracting 2D image features. These features are fed to a graph convolutional network (GCN) that transforms a template mesh to the final output mesh. The graph's topology is defined as per that of the mesh. Each vertex in the graph is initialized with a local feature extracted from the image encoder feature maps. This extraction is performed using a DensePose \cite{guler2018densepose} correspondence map that maps pixel locations to the mesh's visible surface.  


We use a ResNet50 \cite{he2016deep} network that we denote by $f$ to build convolutional image feature maps from an input image $\mathrm{I}$, extracted in a coarse-to-fine fashion at 5 network stages:
$$\{\mathrm{F}_\mathrm{I}^l\}_{1 \leq l \leq 5} = f(\mathrm{I}),$$
$\mathrm{F}_\mathrm{I}^l$ being the feature map at stage $l$. The feature maps' spatial dimensions decrease gradually from the input image dimensions $(H,W)$ downwards and their respective feature dimensions (spatial resolutions) are 64 ($112\times112$), 256 ($56\times56$), 512 $(28\times28)$, 1024 $(14\times14)$ and 2048 ($1\times1$), summing up to $D = 3904$. The final feature is a globally pooled one.

\begin{figure}[t]
\centering
\includegraphics[width=0.9\linewidth]{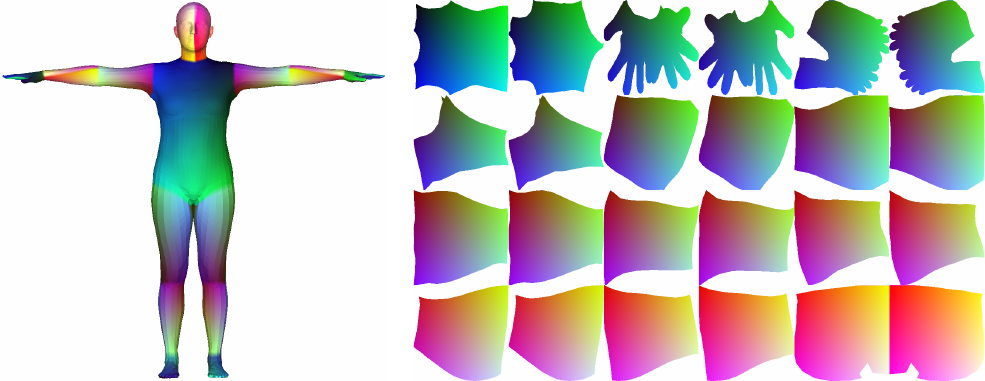}
\caption{Illustration of the template geometry IUV mapping using $24$ body parts. The red channel is representative of the body part, while the green and blue ones span the UV coordinates within each individual part.}
\label{fig:iuv}
\end{figure}

In parallel, the DensePose \cite{guler2018densepose} network, denoted by $h$, predicts a dense mapping from the image pixels to the template mesh, in the form of a 3-channel image $\mathrm{IUV}$: 
$$\mathrm{IUV} = h(\mathrm{I}),$$
where $\mathrm{IUV} \in \llbracket 0,24 \rrbracket \times [0,1]\times [0,1]^{H\times W}$. The first channel indicates which one of 24 pre-defined body parts the pixel belongs to, 0 being the background label. The second and third channels indicate the UV-coordinate of the pixel in a pre-defined UV-map of that body part's template mesh (See Fig.\ref{fig:iuv}).  

Given a vertex $k$ in the template mesh, the corresponding pixel location $c_k \in \llbracket 1,H \rrbracket \times \llbracket 1,W \rrbracket $ in the input image is obtained using the following thresholded nearest-neighbor strategy:
$$\hat{c}_k=\argmin_{(i,j)}\left(||\mathrm{IUV}(i,j) - \overbar{\mathrm{IUV}}_k||_2\right),$$
\[c_k= 
\begin{dcases}
    \hat{c}_k,& \text{if } ||\mathrm{IUV}(\hat{c}_k) - \overbar{\mathrm{IUV}}_k||_2 \leq \delta_k\\
    \emptyset,              & \text{otherwise}
\end{dcases}\]
where $\overbar{\mathrm{IUV}}_k$ is the $k^{th}$ vertex IUV coordinates in the pre-defined template geometry body partitioning and per part UV-map (Fig.\ref{fig:iuv}). Threshold $\delta_k$ is defined as the distance of the $k^{th}$ vertex to its closest adjacent neighbor in the template UV-map space. This thresholding ensures no pixels are assigned to occluded surface regions.

We then construct the input local mesh-borne feature $\mathrm{F}_\mathrm{M}(k)$ for the $k^{th}$ vertex as follows:
$$\mathrm{F}_\mathrm{M}(k)=\left[ \mathrm{F}_\mathrm{I}^1(c_k),\hdots,\mathrm{F}_\mathrm{I}^4(c_k),\mathrm{F}_\mathrm{I}^5,c_k, \overbar{x}_k, \overbar{y}_k, \overbar{z}_k \right]$$
where $\mathrm{F}_\mathrm{I}^l(\emptyset) = \text{0}$ for $1 \leq l \leq 4$ so the non-visible vertices are assigned null local features. We note that the last feature $\mathrm{F}_\mathrm{I}^5$ is a global one and hence does not depend on the spatial 2D sampling. $(\overbar{x}_k, \overbar{y}_k, \overbar{z}_k)$ are the $k^{th}$ vertex coordinates in the T-pose initial mesh, hence $\mathrm{F}_\mathrm{M}(k) \in {\Bbb R}^{D+5}$.

Finally, a GCN $g$, whose topology is defined by the template mesh connectivity, takes as input the mesh local features $\mathrm{F}_\mathrm{M} \in {\Bbb R}^{N_v\times(D+5)}$ and predicts the final mesh vertex coordinates $\mathrm{M} \in {\Bbb R}^{N_v\times 3}$:
$$\mathrm{M}=g(\mathrm{F}_\mathrm{M}),$$
$N_v$ being the total number of vertices ($N_v = 6890$). $g$ uses the formulation from \cite{kipf2016semi} and follows the architecture described in \cite{kolotouros2019convolutional}, where regular graph convolutions are substituted by the semantic graph convolutions introduced in \cite{zhao2019semantic}. We note that body joints $\mathrm{J}$ can be obtained from meshes using a fixed linear regressor $W$: $\mathrm{J}=W\mathrm{M}$, where $\mathrm{J} \in {\Bbb R}^{N_j\times 3}$, $N_j$ being the total number of joints ($N_j = 29$).  

We train the parameters of the convolutional networks $f$ and $g$ following the losses, training scheme, data augmentation and 3D supervision introduced in \cite{moon2020i2l}. Our loss combines $L1$ reconstruction errors between the prediction and the ground-truth for mesh vertices, joints and edges as well as an additional surface normal based constraint:

$$L_{\text{vertex}} = \sum_{i \in \text{vertices}}||v_i - \tilde{v}_i||_1,$$
$$L_{\text{joint}} = \sum_{i \in \text{joints}}||j_i - \tilde{j}_i||_1,$$
$$L_{\text{edge}} = \sum_{(i,j) \in \text{edges}}\left| ||v_i - v_j||_2 - ||\tilde{v}_i - \tilde{v}_j||_2 \right|,$$
$$L_{\text{normal}} = \sum_{k\in \text{faces}} \sum_{(i,j)\in \text{k}}\left|\langle \frac{v_i-v_j}{||v_i-v_j||_2},\tilde{n}_k\rangle \right|,$$

where $\mathrm{J}=[j_1,\dots,j_{N_j}]$ are the predicted 3D joints and $\tilde{\mathrm{J}}=[\tilde{j}_1,\dots,\tilde{j}_{N_j}]$ the ground-truth ones. $\mathrm{M}=[v_1,\dots,v_{N_v}]$ are the predicted mesh 3D vertices while $\tilde{\mathrm{M}}=[\tilde{v}_1,\dots,\tilde{v}_{N_v}]$ are the ground-truth ones. $\tilde{n}_k$ is the normal of the $k^{\text{th}}$ face in the ground-truth mesh $\tilde{\mathrm{M}}$.
The normal and edge losses are used to ensure smoother and more visually pleasing results. The individual losses are combined with the following weighting scheme:
$$L = L_{\text{vertex}} + L_{\text{joint}} + 0.1 L_{\text{normal}} + 0.1 L_{\text{edge}}.$$

$$
$$

\section{RESULTS}

We present in this section our experimental setup in addition to our results. We train our network on datasets Human3.6M \cite{ionescu2013human3} and MSCOCO \cite{lin2014microsoft} following the data augmentation and 3D supervision described in \cite{moon2020i2l}. We use the Adam optimizer \cite{kingma2014adam} in PyTorch on a Quadro RTX 5000 GPU for 12 epochs with a learning rate of $10^{-4}$, followed by another 2 epochs using a learning rate of $10^{-5}$ and finally 1 extra epoch using a learning rate of $10^{-6}$. The image feature extraction network $f$ is initialized with the ImageNet \cite{deng2009imagenet} pre-trained weights.

\subsection{Datasets and evaluation metrics}
\noindent{\bf Human3.6M.} Human3.6M \cite{ionescu2013human3} is a large scale indoor dataset with 3D joint coordinate annotations, and includes multiple subjects performing a variety of actions like walking, sitting and eating. Due to licensing issues, the corresponding groundtruth 3D meshes are not available. Hence, following \cite{choi2020pose2mesh,moon2020i2l}, we use the provided pseudo groundtruth 3D meshes obtained using SMPLify-X \cite{pavlakos2019expressive} for training. However, during inference, we use the groundtruth 3D joint coordinate annotations provided in Human3.6M \cite{ionescu2013human3} to keep evaluation fair. We follow the experiment setting of \cite{choi2020pose2mesh,moon2020i2l} and train our models using subjects S1, S5, S6, S7 and S8. We test the models using subjects S9 and S11.\\
{\bf MSCOCO.} MSCOCO \cite{lin2014microsoft} contains large-scale in-the-wild images with 2D bounding box and human joint coordinates annotations. Following \cite{choi2020pose2mesh,moon2020i2l}, we use the provided pseudo groundtruth 3D meshes obtained by fitting SMPLify-X \cite{pavlakos2019expressive} on the groundtruth 2D poses for training.\\
{\bf 3DPW.} 3DPW \cite{von2018recovering} is also an in-the-wild dataset consisting of 60 video sequences captured mostly in outdoor conditions. It contains 3D body pose and mesh annotations. We use this dataset only for evaluation purposes using its test set following \cite{choi2020pose2mesh,moon2020i2l}.

Concerning evaluation metrics, we report our performance for 3D pose estimation, in line with seminal work, using two metrics, namely mean per joint position error (MPJPE) and mean per joint position error after procrustes analysis (PA-MPJPE). MPJPE is the Euclidean distance (in mm) between the predicted and groundtruth 3D joints after root joint alignment. PA-MPJPE is the same after a further rigid alignment using Procrustes analysis.  

\subsection{Comparison with state-of-the-art methods}

We evaluate our contribution numerically using the 3DPW \cite{von2018recovering} and Human3.6M \cite{ionescu2013human3} datasets. We report the MPJPE and PA-MPJPE metrics of our method in Table \ref{tab:table1} when using only one global feature for all vertices ($\mathrm{F}_\mathrm{I}^5$) \textit{Global (Ours)}, and also when using vertex specific local features (full $\mathrm{F}_\mathrm{M}$) \textit{Local (Ours)}. For a fair comparison with the competition, we show other methods (\cite{kanazawa2018end,kolotouros2019convolutional,kolotouros2019learning,choi2020pose2mesh,moon2020i2l}) trained on the same data as us and we relay their performances as they were reported in \cite{moon2020i2l}.

\begin{figure}[t]
\centering
\includegraphics[width=0.49\linewidth]{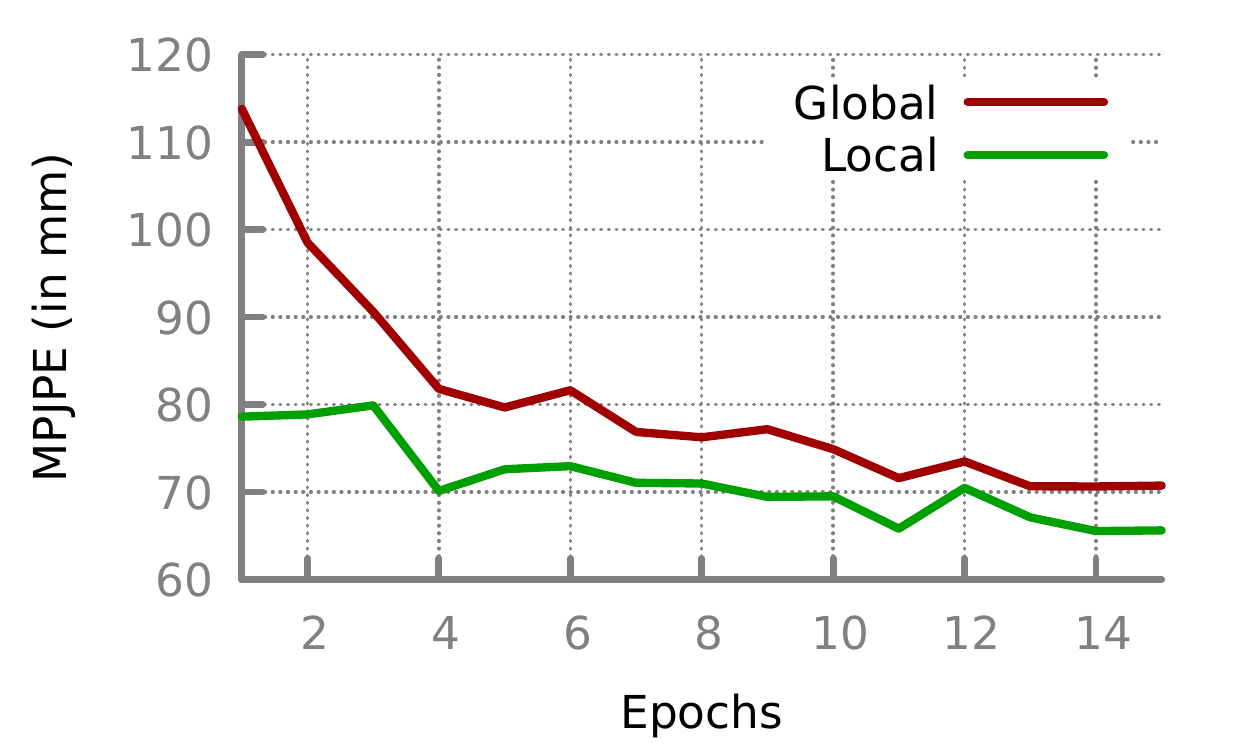}
\includegraphics[width=0.49\linewidth]{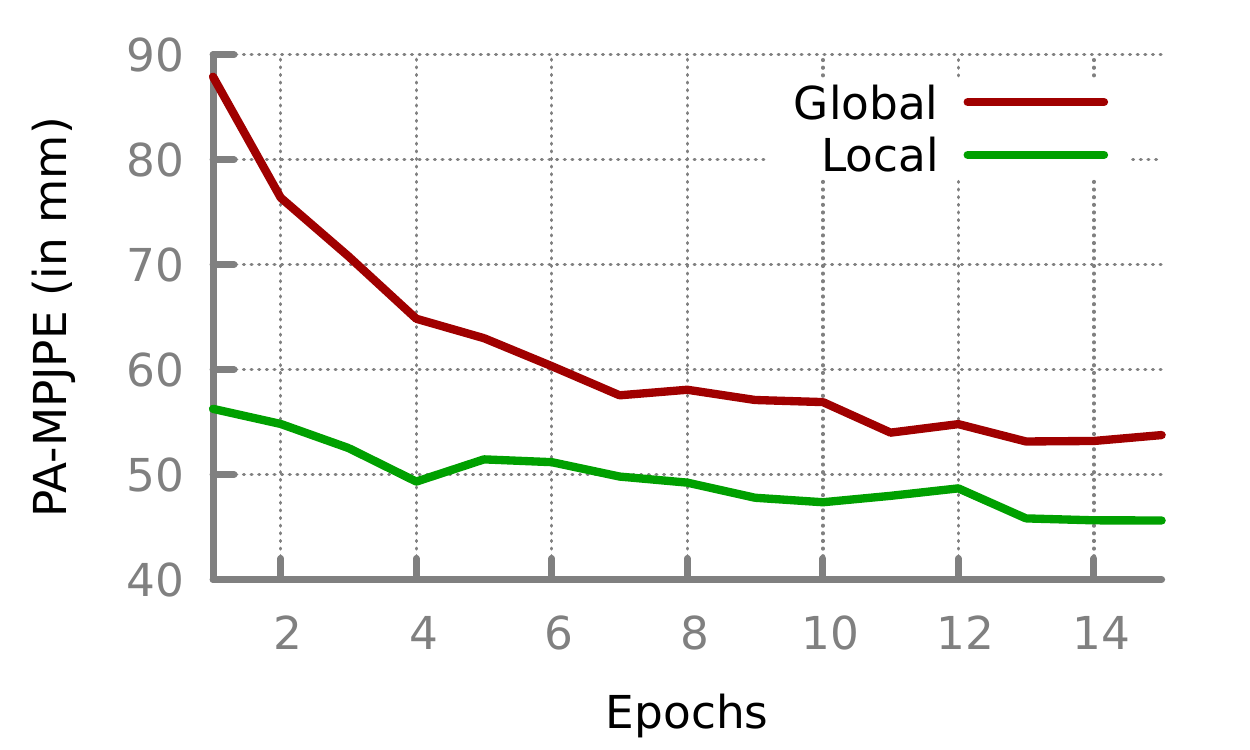}
  \caption{MPJPE \textit{(left)} and PA-MPJPE \textit{(right)} testing losses on Human3.6M \cite{ionescu2013human3} for our baseline using only a global image feature (Global) and our proposed approach using per vertex pixel-aligned local features (Local).}
  \label{fig:comparison_plots}
\end{figure}

\begin{table}[t]
  \centering
\begin{tabular}{ c|cc|cc }
 \hline
 Methods&\multicolumn{2}{|c|}{Human3.6M \cite{ionescu2013human3}}& \multicolumn{2}{|c}{3DPW \cite{von2018recovering}}\\ 
 & MPJPE &PA-MPJPE &MPJPE &PA-MPJPE\\
 \hline
 HMR~\cite{kanazawa2018end} &153.2 &85.5 &300.4 &137.2\\
 GraphCMR~\cite{kolotouros2019convolutional} &78.3 &51.9 &126.5 &80.1\\
 SPIN~\cite{kolotouros2019learning} &72.9 &85.5 &113.1 &71.7\\
 Pose2Mesh~\cite{choi2020pose2mesh} &67.9 &49.9 &91.4 &60.1\\
 I2L-MeshNet~\cite{moon2020i2l} &55.7 &41.7 &95.4 &60.8\\
 \hline
 Global (Ours) &70.74 &53.76 &112.51 &67.98\\
 \textbf{Local} (Ours) &65.61 &45.62 &110.31 &66.52\\
 \hline
\end{tabular}
\caption{Comparison of MPJPE and PA-MPJPE on Human3.6M \cite{ionescu2013human3} and 3DPW \cite{von2018recovering}. All
methods are trained on Human3.6M \cite{ionescu2013human3} and MSCOCO \cite{lin2014microsoft}.} \label{tab:table1}
\end{table}


We firstly showcase the effect of using local features on the convergence of our model in Fig. \ref{fig:comparison_plots}. For both of the PA-MPJPE and MPJPE metrics, our pixel-aligned mesh features enable the model to reach significantly lower generalization errors on Human3.6M \cite{ionescu2013human3} at the same training epoch compared to our global feature baseline. Furthermore, we note that our \textit{Global} baseline shares the same network design as GraphCMR \cite{kolotouros2019convolutional}. However, by substituting regular convolutions with learnable adjacency matrix  ones \cite{zhao2019semantic} and training with the same supervision and training scheme as in \cite{moon2020i2l}, we manage to improve its performance by roughly 8mm in MPJPE for Human3.6M \cite{ionescu2013human3}, 14mm in MPJPE and 13mm in PA-MPJPE for 3DPW \cite{von2018recovering}. Our proposed \textit{Local} method improves on our \textit{Global} baseline substantially in almost all figures, by roughly 5mm in MPJPE and 8mm in PA-MPJPE for Human3.6M \cite{ionescu2013human3}, and 2mm in MPJPE for 3DPW \cite{von2018recovering}. It is noteworthy that our \textit{Local} version also achieves competitive results in comparison to the state-of-the-art, as it outperforms all methods presented in the table on Human3.6M \cite{ionescu2013human3} except for  I2L-MeshNet~\cite{moon2020i2l}, while ranking $3^{rd}$ on 3DPW \cite{von2018recovering} closely behind I2L-MeshNet~\cite{moon2020i2l} and Pose2Mesh~\cite{choi2020pose2mesh}, which uses 2D joints (from HRNet \cite{sun2019deep}) as input rather than a RGB image. While \cite{choi2020pose2mesh} is advantaged by the 2D pose input, we believe the performance of \cite{moon2020i2l} is by virtue of their Lixel architecture which is not readily applicable to irregular graphs.       

\begin{figure}[h!]
\centering
\includegraphics[width=\linewidth]{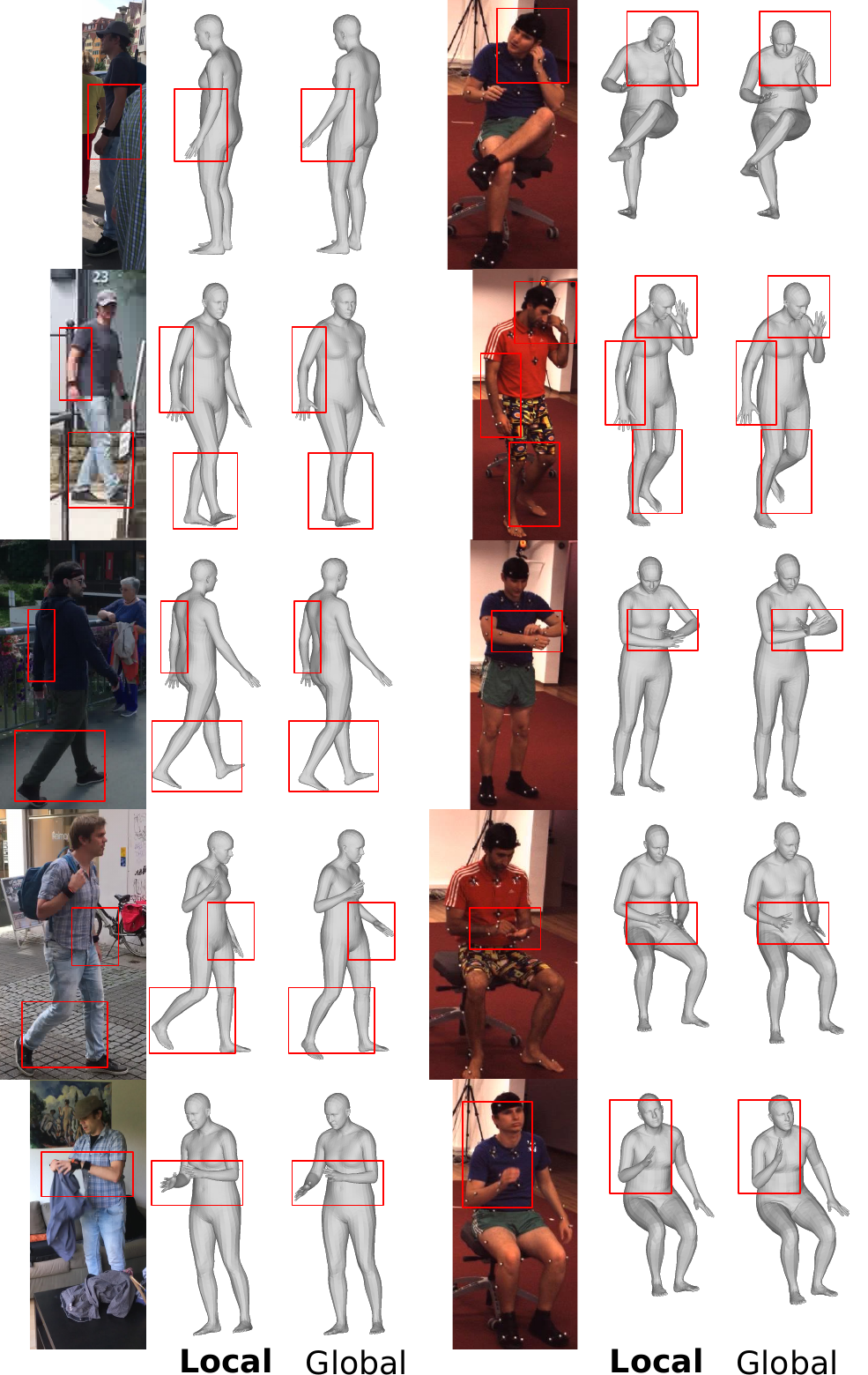}
\caption{Comparison of our baseline using only a global image feature (Global) and our proposed approach using per vertex pixel-aligned local features (Local) on the 3DPW \cite{von2018recovering} \textit{(left)} and Human3.6M \cite{ionescu2013human3} \textit{(right)} datasets.
}
\label{fig:qual}
\end{figure}

The numerical superiority of our contribution compared to our baseline is also confirmed with qualitative results. As shown in Fig.\ref{fig:qual}, we notice that using local image features in the GCN stage yields improved visual results, as witnessed by these examples from the 3DPW \cite{von2018recovering} and Human3.6M \cite{ionescu2013human3} datasets. The red boxes in the figure illustrate in particular the better positioning of the body limbs with our \textit{Local} method compared to our \textit{Global} baseline. We note that following \cite{kolotouros2019convolutional}, we show the results after an additional linear layer trained to predict SMPL instances from the previous output for smoother visual results.

\section{CONCLUSION}

We presented a method for 3D human shape and pose estimation from a single RGB image. The method is model-free and relies on a GCN that starts from a template T-posed mesh and regresses the final vertex coordinates. Contrarily to seminal work \cite{kolotouros2019convolutional}, we propose to initialize the graph convolutions with pixel-aligned vertex-specific features instead of only one global feature. These features are extracted at multiple feature map stages of an image CNN, and mapped subsequently to the graph vertices using a pixel-to-surface correspondence map \cite{guler2018densepose}. Our results demonstrated the benefit of using local features in GCN based human 3D shape and pose estimation. Next, we will attempt to make the entire pipeline fully differentiable, by including the correspondence estimation network $h$ training in the end-to-end learning framework alongside the image feature network $f$ and graph network $g$.      


{\small
\bibliographystyle{ieee}
\bibliography{egbib}
}

\end{document}